\documentclass[11pt,en]{elegantpaper}
\usepackage{amsmath,amssymb,amsfonts}
\usepackage{algorithmic}
\usepackage{graphicx}
\usepackage{textcomp}
\usepackage{wrapfig}

\title{GITSR: Graph Interaction Transformer-based Scene Representation for Multi Vehicle Collaborative Decision-making}
\author{Xingyu Hu, Lijun Zhang\textsuperscript{*}, Dejian Meng, Ye Han\textsuperscript{*}, and Lisha Yuan}

\usepackage{array}

\begin{document}

\maketitle

\footnote{Xingyu Hu: 2410254@tongji.edu.cn, Lijun Zhang: tjedu\_zhanglijun@tongji.edu.cn, Dejian Meng: mengdejian@tongji.edu.cn, Ye Han: hanye\_leohancnjs@tongji.edu.cn, Lisha Yuan: 2051922@tongji.edu.cn.}

\begin{abstract}
In this study, we propose GITSR, an effective framework for \underline{G}raph \underline{I}nteraction \underline{T}ransformer-based \underline{S}cene \underline{R}epresentation for multi-vehicle collaborative decision-making in intelligent transportation system. In the context of mixed traffic where Connected Automated Vehicles (CAVs) and Human Driving Vehicles (HDVs) coexist, in order to enhance the understanding of the environment by CAVs to improve decision-making capabilities, this framework focuses on efficient scene representation and the modeling of spatial interaction behaviors of traffic states. We first extract features of the driving environment based on the background of intelligent networking. Subsequently, the local scene representation, which is based on the agent-centric and dynamic occupation grid, is calculated by the Transformer module. Besides, feasible region of the map is captured through the multi-head attention mechanism to reduce the collision of vehicles. Notably, spatial interaction behaviors, based on motion information, are modeled as graph structures and extracted via Graph Neural Network (GNN). Ultimately, the collaborative decision-making among multiple vehicles is formulated as a Markov Decision Process (MDP), with driving actions output by Reinforcement Learning (RL) algorithms. Our algorithmic validation is executed within the extremely challenging scenario of highway off-ramp task, thereby substantiating the superiority of agent-centric approach to scene representation. Simulation results demonstrate that the GITSR method can not only effectively capture scene representation but also extract spatial interaction data, outperforming the baseline method across various comparative metrics.
\end{abstract}

\section{Introduction}
\label{sec:introduction}
Autonomous vehicles have garnered significant research attention over the past two decades, driven by their substantial potential for societal and economical advancement. The efficient coordination of driving decisions among CAVs promises not only to enhance safety and operational efficiency but also to reduce energy consumption [1]. However, in dynamic traffic scenarios, the intricate interplay between scenarios and traffic participants presents formidable challenges for CAVs in making decisions that are safe, efficient, and comfortable [2]. The Internet of Vehicles (IoV) technology integrates Vehicle-to-Vehicle (V2V) and Vehicle-to-Infrastructure (V2I) communications with artificial intelligence (AI) to offer innovative solutions for CAVs to process dynamic traffic scene information and perform collaborative driving decisions [3]. In this context, methods based on deep reinforcement learning (DRL) are becoming more and more popular because the intelligent agent can continuously learn through interaction with the driving environment, extract environmental information through deep learning, and make decisions through reinforcement learning algorithms [4]. However, modeling and representing scene information effectively, processing and calculating it to adapt to various complex traffic environments, while achieving high-quality collaborative decision-making in real-time dynamic settings, has emerged as a formidable research challenge. Concurrently, the research on autonomous vehicle decision-making is increasingly focusing on more complex scenarios. The crux of the challenge lies in the representation of the state, which must encompass the elements, characteristics, and interactions in the dynamic scene. Addressing this will become one of the key issues of the DRL methods [5].

To this end, we introduce GITSR, a novel graph interaction Transformer-based scene representation framework for multi-vehicle collaborative decision-making. This framework leverages the Transformer architecture to capture scene information and employs a graph structure to model spatial interaction, thereby enhancing the multi-vehicle collaborative decision-making ability of reinforcement learning. Firstly, we extract features from the dynamic driving environment within the context of intelligent networking, meticulously considering both the local interaction and global communication attributes of CAVs. We perform local reconstruction reasoning on scene input information, introduce the Transformer module to process information and enhance understanding of surrounding traffic scene for CAVs. We conduct local reconstruction reasoning on the input scene information and introduce the Transformer module to process this data, thereby enhancing the CAVs’ comprehension of the surrounding traffic environment. Then, we represent the dynamic traffic scene as a graph, based on global communication attributes, and introduce GNN to extract spatial interaction features. This approach is advantageous as it optimally utilizes the information from all CAVs within dynamic traffic scenarios. It aids CAVs in scene comprehension and the transmission of upstream and downstream information. Moreover, it establishes the spatial interaction dynamics of the traffic environment, optimizing the collaborative driving decision-making capabilities. The main contributions of this article can be summarized as follows:

\begin{itemize}
  \item [1)]
  A collaborative decision-making framework for intelligent connected vehicles that integrates Transformer and GNN is designed, which is tailored for scene extraction and interaction modeling from the perspective of state representation, thus significantly enhancing the state representation to improve the reinforcement learning effect.
  \item [2)]
  A local representation method based on Transformer to reconstruct reasoning from scene features is proposed. This method reconstructs the scene representation with a focus on all CAVs and employs GNN to extract spatial interaction behaviors between the motion information of traffic participants. The GITSR framework can make full use of the information extracted from features to assist all CAVs in comprehending both local scene details and global interaction dynamics.
  \item [3)]
  The framework is verified in a challenging interactive collaborative driving environment. The results show that GITSR has advantages over advanced algorithms in terms of safety, efficiency, and task success rate. At the same time, we have conducted an assessment of the influence of various components within the GITSR framework on its overall performance.
\end{itemize}

\section{Background and related work}
\label{sec:background and related work}

\subsection{DRL for Autonomous Driving Decision-making}
There are primarily two approaches to decision-making for autonomous vehicles: rule-based and learning-based. The majority of decision modules in Baidu Apollo are rule-based, characterized by their simplicity of implementation and the clarity of their logic, which is derived from manually formulated rules [6]. However, as traffic scenarios grow increasingly complex, this approach becomes less efficient and challenging to apply. DRL amalgamates deep learning with reinforcement learning, enabling self-learning through environmental interactions without predefined complex rules. It can handle high-dimensional and complex decision-making problems and become one of the mainstream methods for autonomous vehicles behavior decision-making [7]. Especially in recent years, with the advancement of deep learning, a large number of cutting-edge algorithms that integrate reinforcement learning have yielded remarkable outcomes [8], [9], [10], [11]. Inspired by Natural Language Processing (NLP), autonomous vehicles can better select actions by remembering some history, and learn the long-term correlation between scenes and motion states through long short-term memory (LSTM) [12]. The attention mechanism enables neural networks to discover interdependencies in a variable number of inputs. Leurent et al. [13] designed an attention mechanism based on the scene of a non-signal intersection to successfully learn to identify and utilize the interaction mode of controlling nearby traffic, and realized the visualization of the attention matrix. Li et al. [14] have successfully integrated separable convolution with the Transformer architecture for vehicle lane-changing scenarios, resulting in a lightweight yet high-performing solution. Building on this, our research applies the DRL method to the behavioral decision-making process of autonomous vehicles and extends its application to multi-vehicle collaborative decision-making in higher-dimensional contexts.

\begin{figure}[!t]
\centerline{\includegraphics[width=\columnwidth]{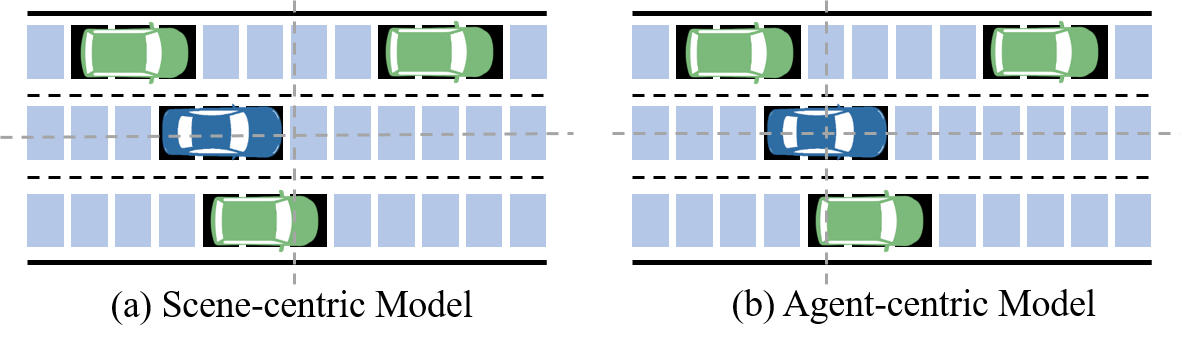}}
\caption{Scene-centric model represents the state of traffic participants in a unified coordinate system. 
Agent-centric model uses the agent of interest as the center of the coordinate system, and other traffic participants are represented relative to the agent.}
\label{fig1}
\end{figure}

\begin{figure*}[!t]
\centerline{\includegraphics[width=\textwidth]{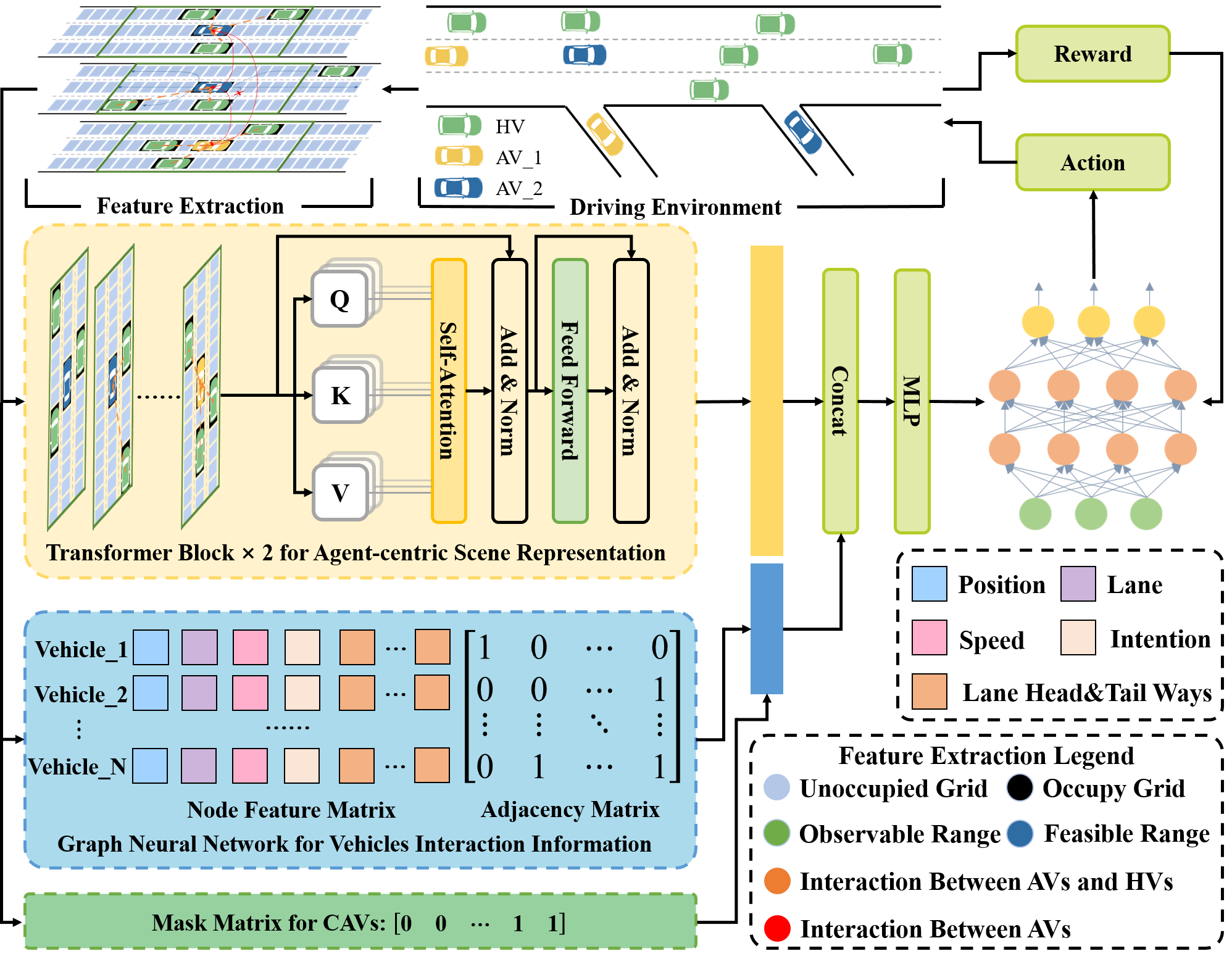}}
\caption{An overview of our GITSR framework. After extracting driving scene features, the Transformer module encodes the scene information, and the GNN constructs the motion information as a spatial interaction graph. The reinforcement learning module calculates the output decision behaviors.}
\label{fig2}
\end{figure*}

\subsection{Scene Representation in DRL}
The realm of DRL-based behavior decision-making for autonomous vehicles continues to face numerous challenges. A key issue is the accurate representation of traffic scenes, including both static and dynamic road elements as well as the status of traffic participants [15]. Traffic participants are highly interactive in real time, which significantly influence the interpretation of the traffic scene and the output of decision-making behaviors of CAVs. Over the past few years, addressing the challenge of scene representation has emerged as a central focus in a multitude of studies, which can include vehicle status information, vehicle-observed environmental information, and the interactions among traffic participants. The feature list method represents the status of CAVs and surrounding vehicles observed by them, such as position, speed, and heading, in a matrix list. The encoding approach using motion information has been widely used in research [16], [17], [18]. However, a limitation of this method is that the number of selected surrounding vehicles and the ordering of the list directly impact the outcome, making it challenging to adapt to scenes that fluctuate dynamically. A common way to overcome this limitation is to employ a spatial grid representation, construct the scene into a grid, and no longer select surrounding vehicles for status representation, but instead cover them with occupied spatial grid. A pivotal aspect of the spatial grid method is the selection of the coordinate system, which typically falls into two categories: scene-centric and agent-centric models [19]. The scene-centric model depicts the status of traffic participants in a unified coordinate system after anchoring the scene, usually by discretizing the entire scene into a spatial grid akin to an aerial map. For example, Y. Zheng et al. [20] mapped the entire urban area into a spatial grid and developed a method to represent all vehicles within a coordinate system.
Differently, the agent-centric model [1], [21] uses the CAV of interest as the central coordinate, and the surrounding traffic participants are represented by their states relative to the vehicle, which can be regarded as scene-centric reconstruction reasoning, as shown in Fig. 1. 
In our study, we evaluate the performance of these two models in decision-making tasks and employ a more powerful Transformer encoding structure. In addition, we preserve the motion information from the feature list method and integrate GNN framework to model the spatial interactions among multiple vehicles, thereby improving the decision-making performance.

\section{Methodology}
\label{sec:methodology}
In this research, our proposed GITSR framework for DRL is designed to focus on the effective scene representation and interaction modeling of CAVs within dynamic mixed traffic environments to improve collaborative decision-making capabilities. This section mainly introduces the overall framework of GITSR and its details, as shown in Fig. 2. After feature extraction of the driving environment, it is divided into three parts: scene representation, interactive behaviors modeling and mask matrix. The scene representation is input to the Transformer module for encoding and calculation to extract map information. The interactive behaviors are represented as a state space matrix and an adjacency matrix through graph neural network. Mask matrix is used to filter out non-autonomous vehicles information. Ultimately, the RL module synthesizes and processes the scene representation and interaction behaviors, then outputs the determination of driving actions. Once the CAVs execute these actions, the environment feedback rewards to facilitate the updating of the network.

\subsection{Problem formulation}
Based on the background of intelligent networking, we propose the problem of multi-vehicle cooperative decision-making in mixed traffic environments. The multi-vehicle cooperative decision-making problem based on RL can be formulated as a MDP, which can be represented as a tuple  ($s_t,a_t,r_t,s_{t+1},\gamma$). In the autonomous driving scenario, each CAV can only obtain environmental information within its perception range due to sensor constraints, and must communicate with each other to share information. Therefore, at each time step, CAVs construct a state representation $s_t \in S$  of through information sharing, and each CAV executes an action $a_t \in A$ that causes the environment to transfer to state $s_{t+1}$. Ultimately, all CAVs share a reward function $r_t$. The process is then repeated, with the goal of allowing CAVs to choose actions at each time step to maximize their expected future discounted reward $E$[$\sum_{t=0}^{\infty}$$\gamma^tr_t$], where $r_t$ is the reward obtained at time $t$. The discount factor $\gamma$ determines how much immediate rewards are favored over more distant rewards.

\subsection{Scene representation with Transformer}
We implement an agent-centric scene representation approach. Specifically, we first rasterize the input scene to extract scene features. We assume that each CAV can perceive the traffic environment within a 50-meter radius to the front and rear of the vehicle, and reconstruct the local traffic scene with the vehicle as the center coordinate of the grid to obtain the local map $map_i \in \mathbb{R}^{n_{lanes} \times 51}$ of the $i$-th CAV. The ego vehicle occupies $map_i$[$j$, $centric$] according to the lane $j$ it is in, and the other vehicles in the perception range occupy $map_i$ according to their relative positions and lanes. Notably, we employ a grid occupation method based on vehicle speed to more accurately represent the traffic scene. All unoccupied local grids are designated with a value of 0. Fig. 3. shows a schematic diagram of our scene representation method. In addition, all CAVs can engage in communication to share their individual local traffic scene information, thereby assembling a comprehensive multi-vehicle scene representation.

After extracting the driving scene to generate the multi-vehicle scene representation, encoding it in the GITSR framework should have the following two properties: 1) The algorithm should be able to capture the interactive information between the vehicle and the surrounding traffic scene; 2) The algorithm should be able to effectively extract the shared information of multi-vehicle communication. Therefore, deploying the Transformer algorithm for information extraction in the GTISR framework is an effective method, as the Transformer can focus on key information among a large amount of input information and ignore unimportant information. In the scene representation, the Transformer can achieve: 1) Capturing the local dynamic changes of each CAV and analyzing the feasible domain; 2) Utilizing the shared information among all CAVs to effectively guide the formulation of collaborative decision-making and driving actions.

Since Transformer was first proposed by Vaswani et al. [22] in 2017, it has been widely used in NLP and Computer Vision (CV) fields [23], [24]. We will first introduce the core Multi-Head Attention (MHA) mechanism of Transformer Block. In the self-attention mechanism of MHA, the input information is passed through to obtain the embedding vector $X$. Then, $X$ is multiplied with three different weight matrices $W_q$, $W_k$ and $W_v$ to obtain three different vectors $Q$, $K$ and $V$, which represent the query, key and value respectively. The computational formula is as follows:
\begin{equation}
\begin{aligned}
Q=XW_q\\
K=XW_k\\
V=XW_v
\label{eq}
\end{aligned}
\end{equation}
where the dimensions of the three weight matrices are the same.

\begin{figure}[!t]
\centerline{\includegraphics[width=\columnwidth]{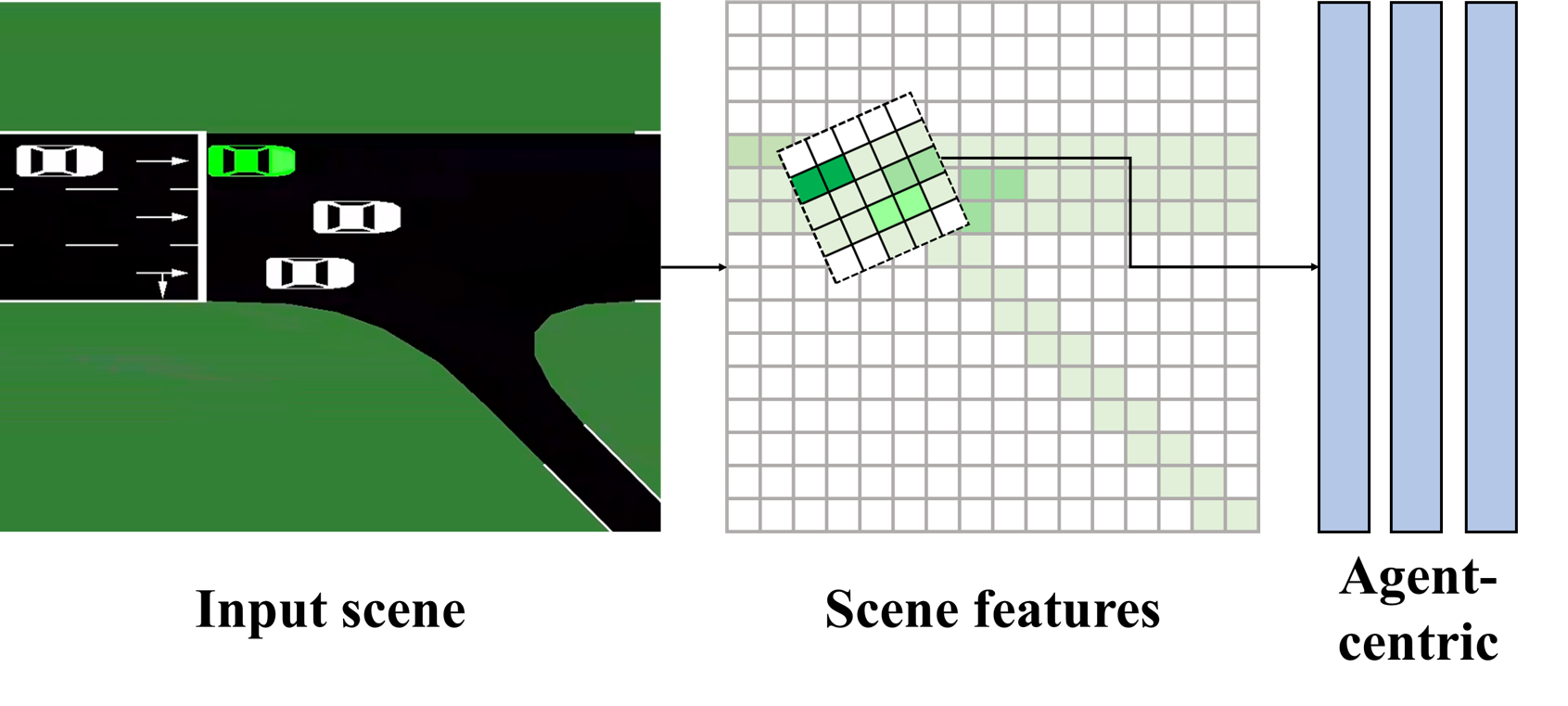}}
\caption{Schematic diagram of the scene representation method. First, the entire input scene is rasterized to obtain scene features. The relative information of the surrounding traffic participants is reconstructed and inferred using the center of each CAV as the coordinate system. Finally, an agent-centric scene representation is obtained, where the darker the grid, the faster the speed.}
\label{fig3}
\end{figure}

The attention matrix is obtained by scaling the dot product of the reciprocal square root of the number of columns $\sqrt{d_k}$ of the $Q$ and $K$ matrices and normalizing it:
\begin{equation}\text{Attention}(Q,K,V)=\text{softmax}(\frac{QK^T}{\sqrt{d_k}})V.\label{eq}\end{equation}

A layer of Transformer Block contains $h$ attention modules that calculate the self-attention matrix in parallel and then concatenate the results:
\begin{equation}
\text{MHA}(X)=\text{concat}(head_1,...,head_h)W^o
\label{eq}
\end{equation}
where $head_i=\text{Attention}(XW_q,XW_k,XW_v)$, $W_i^q \in \mathbb{R}^{d_{model} \times d_q}$, $W_i^k \in \mathbb{R}^{d_{model} \times d_k}$, $W_i^v \in \mathbb{R}^{d_{model} \times d_v}$, $W^o \in \mathbb{R}^{d_{model} \times hd_v}$.

We concatenate the map obtained by the $i$-th CAV row by row to obtain the scene representation matrix $SR_i \in \mathbb{R}^{1 \times {153}}$, and all $m$ CAVs in the scene get the model input $SR=$\{$SR_1;SR_2;...;SR_m$\}. Since each $SR_i$ has the same dimension, an embedding layer is used to encode the embedding vector of the same length that can be further processed by the Transformer. The L-layer Transformer Blocks are composed of alternating MHA and Multi-Layer Perceptron (MLP). A LayerNorm (LN) layer is added before each block, and after each MHA and MLP, the features of the previous layer are added to the output to merge and retain the original features. The overall calculation is as follows:

\begin{equation}
\begin{aligned}
&X_0=\text{Embedding}[SR_1;SR_2;...;SR_m]\\
&X_l^{\prime}=\text{MHA}(X_{l-1}),l=1,...,L\\
&X_l=\text{LN}(\text{MLP}(X_l^{\prime}+X_{l-1})),l=1,...,L\\
&y=X_L
.\label{eq}
\end{aligned}
\end{equation}

In general, the self-attention mechanism of Transformer encoder can help each CAV capture the traffic scene information around the ego vehicle, and MHA enables all CAVs to collaboratively process shared information to guide driving actions generation.

\subsection{Spatial interactive behaviors with GNN}
Dynamic traffic scenes also have spatial interaction characteristics, that is, the distribution and relationship of vehicles and their behaviors in space, which together affect the dynamic changes of traffic flow. In order to effectively represent this characteristic, after extracting the motion information of the vehicles, we construct the dynamic traffic scene as a graph. Specifically, the modeled spatial interaction behaviors is represented by $G=(N,E)$. $N=\{n_1,n_2,...,n_{|n|}\}$ represents the set of all vehicle features, and $E=\{e_1,e_2,...,e_{|\varepsilon|}\}$ represents the set of interaction relationships between them. $|n|$ constitutes the total number of vehicles in traffic, and $|\varepsilon|$ represents the total number of vehicle interaction relationships.

The feature matrix $N$ represents the longitudinal position, speed, lane, vehicle category, lane head and tail ways of each vehicle in the scene. Therefore, the feature matrix can be expressed as:
\begin{equation}
N=
\begin{gathered}
\begin{bmatrix}
  X_1 & V_1 & L_1 & I_1 & H_1 & T_1 \\
  X_2 & V_2 & L_2 & I_2 & H_2 & T_2 \\
  \multicolumn{6}{c}{\cdots} \\
  X_i & V_i & L_i & I_i & H_i & T_i \\
  \multicolumn{6}{c}{\cdots} \\
  X_n & V_n & L_n & I_n & H_n & T_n 
\end{bmatrix}
\end{gathered}
.\label{eq}
\end{equation}

The motion information of the   vehicle in the scene can be expressed as follows:

\begin{equation}
\left\{
\begin{array}{lr}
X_i=x_{i\_position}/x_{road} & \\
V_i=v_{i\_speed}/v_{max} & \\
L_i \in \{1,2,3\} & \\
I_i \in \{1,2,3\} & \\
H_i=[h_1,h_2,\cdots,h_l]/x_{road} & \\
T_i=[t_1,t_2,\cdots,t_l]/x_{road} & \\
\end{array}
\right
.\label{eq}
\end{equation}
where $x_{i\_position}$ is the current longitudinal position of the vehicle, $x_{road}$ is the total length of the road; $v_{i\_speed}$ is the current speed of the vehicle, $v_{max}$ is the maximum speed limit of the road; $L_i$ is the lane that the vehicle is currently in; $I_i$ is the category of the vehicle ($V_i$=1 or 2 indicates a CAV with different driving task, otherwise it is a human-driven vehicle); $H_i$ includes the headways between the $i$-th vehicle and the vehicle immediately ahead of it in all lanes. $T_i$ includes the headways between the $i$-th vehicle and the vehicle immediately behind it in all lanes.

In the context of intelligent networking, we consider the interaction of multiple vehicles in a space, where each CAV is associated with other communicative vehicles in the scene. Specifically, we focus on the interaction between the $i$-th CAV and all vehicles $j$-th in the scene, denoted as $e_{ij} \in \{0,1\}$, where $e_{ij}=1$ means that there is interaction between the $i$-th vehicle and the $j$-th vehicle, otherwise there is no interaction. In order to represent the spatial interaction behaviors, we make the following assumptions: 1) All CAVs can communicate and interact with each other; 2) CAVs can interact with HDVs within their surrounding perception range; 3) CAVs can interact with themselves. Based on the above assumptions, the adjacency matrix $E$ is obtained as:
\begin{equation}
E=
\begin{gathered}
\begin{bmatrix}
  e_{11} & e_{12} & \cdots &  & \cdots & e_{1n} \\
  e_{21} & e_{22} & \cdots &  & \cdots & e_{2n} \\
  \vdots & \vdots & \ddots &  &  & \vdots \\
   &  &  & e_{ij} &  &  \\
  \vdots & \vdots &  &  & \ddots & \vdots \\
  e_{n1} & e_{n2} & \cdots &  & \cdots & e_{nn} \\
\end{bmatrix}
\end{gathered}
.\label{eq}
\end{equation}

In order to make the output dimension of GNN consistent with Transformer, we introduce a mask matrix $M$ to retain the CAVs information and remove the HDVs information. If $M_i=1$, it means the current index is CAV, otherwise it is HDV:
\begin{equation}
M=[m_1,m_2,\cdots,m_i,\cdots,m_n]
.\label{eq}
\end{equation}

\begin{figure}[!t]
\centerline{\includegraphics[width=\columnwidth]{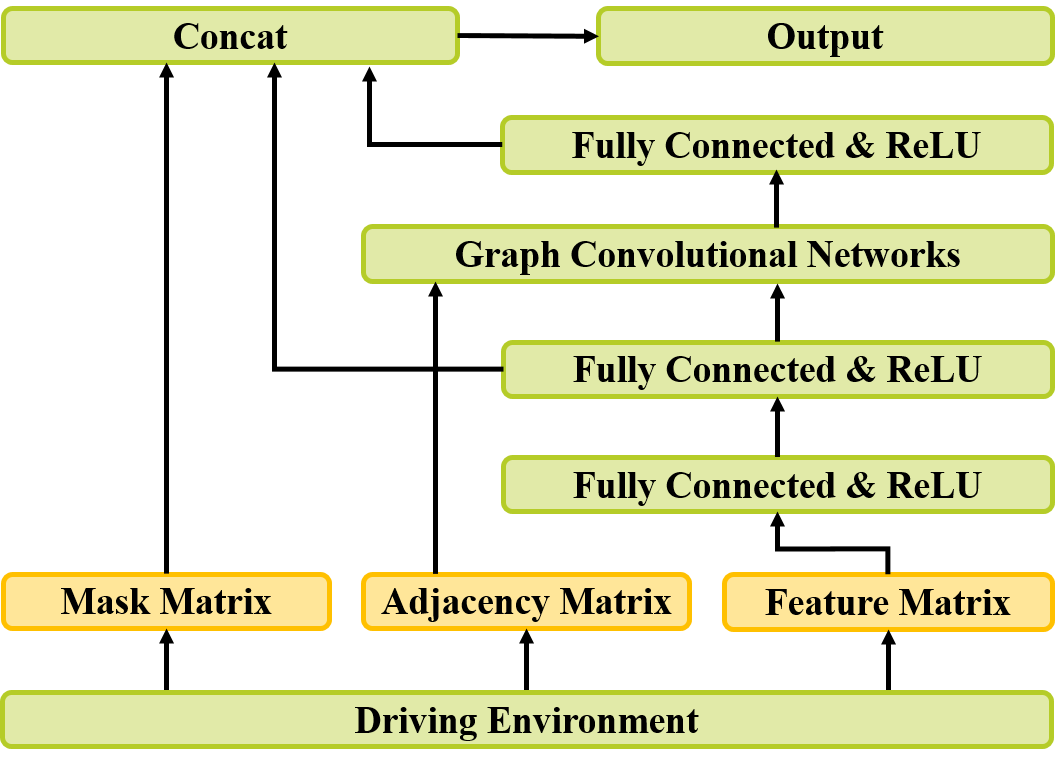}}
\caption{The graph neural network computation framework.}
\label{fig4}
\end{figure}

Graph Convolutional Neural network (GCN) is an algorithm that performs convolution operations directly on graphs [25]. After constructing the dynamic traffic scene as a graph $G=\{N,E\}$, we extract the feature matrix, adjacency matrix and mask matrix, and directly use the neural network to model the entire graph. The calculation framework of GCN is shown in Fig. 4, and the calculation formula is as follows:
\begin{equation}H^{(l+1)}=G(N,E)=\sigma(\widetilde{D}^{-\frac{1}{2}}\widetilde{A}\widetilde{D}^{-\frac{1}{2}}H^{(l)}W^{(l)}).\label{eq}\end{equation}

In the (9), $\widetilde{A}=A+I$, $I$ represents the unit matrix with the same dimension as $A$; $\widetilde{D}=D+I$ , $D$ represents the degree matrix, $D_{ii}=\Sigma_jA_{ij}$ ; $H^{(l)}$ represents the features of each node in the $l$-th layer,   represents the original node features, that is, $H^{(0)}=N$; $W^{(l)}$ represents the learnable parameters of the $l$-th layer. $\sigma$ represents the activation function, and we use ReLU.

\subsection{Decision-making with RL}
As previously discussed, the state space consists of scene representation and motion information. Such a state representation has the following two advantages: 1) the traffic scene can be represented from multiple dimensions; 2) more effective information can be extracted by using the characteristics of different representations. We use Transformer to capture the interaction information between vehicles and scenes, and GNN to extract the spatial interaction behaviors of vehicles. This innovation fully utilizes the information sharing between CAVs by using the characteristics of different neural networks to process complex information in dynamic mixed traffic scenarios. Ultimately, the two information outputs are spliced and input into RL to improve the multi-vehicle collaborative decision-making training process. The formula for the $Q$ value of the driving actions of CAVs generated by RL is as follows:
\begin{equation}Q(s,a)=\phi^{RL}(\text{Concat}(X_L+H_l))\label{eq}\end{equation}
where $Q(s,a)$ represents the $Q$ value of the CAVs driving actions, $\phi^{RL}$ represents the RL policy network, $X_L$ represents the output of the Transformer network, and $H_l$ represents the output of the GNN network.

We employ the classic reinforcement learning algorithm Deep Q-network (DQN) [26] to process the information output by the encoder to output multi-vehicle collaborative driving actions. DQN is a value-based reinforcement learning algorithm that introduces deep neural networks into Q-learning. By processing high-dimensional inputs, it achieves effective decision-making in complex environments. Q-learning is a model-free reinforcement learning algorithm that estimates the expected reward of taking an action in a given state by learning the state-action value function $Q(s,a)$. Q-learning needs to maintain a state-action value table and store a $Q$ value for each possible state-action pair. When the number of states and actions increases, the required storage space and computing time will increase exponentially, limiting the application of Q-learning in high-dimensional environments. DQN approximates $Q(s,a)$ through the function $Q(s,a;\theta)$, solving the problem that traditional Q-learning cannot effectively handle in high-dimensional state space. Specifically, the main idea of DQN is that $Q$ value can be parameterized as $Q(s,a;\theta)$ in the neural network, the state space is used as the input of the neural network, and the action that can obtain the maximum reward is selected as the output action in the action space according to the $Q$ value. In addition, the parameters are updated by sampling from the replay pool and training another target network $Q(s,a;\theta^{\prime})$. The $Q$ value calculation process is as follows:
\begin{equation}y_t^{DQN}=R_{t+1}+\gamma \mathop{\arg\max}\limits_aQ(s,a;\theta^{\prime}).\label{eq}\end{equation}

In our work, we utilize a single DQN to output the actions Q values of all CAVs at the same time, called MADQN [27]. Our objective is to maximize the cumulative reward of each episode, so the reward is the sum of the state values of all CAVs at the current moment.

\section{Experiment}
\label{sec:experiment}

\subsection{Driving Environment}
In order to evaluate the performance of GITSR in the driving environment, we build a challenging highway dual ramp exit scenario based on the FLOW [28] platform, as shown in Fig. 2. In a 400-meters-long highway, there are ramps exit at 250-meter and 370-meter respectively. There are two types of vehicles in the environment, the green cars represent HDVs, and the yellow and blue cars represent CAVs. Both types of vehicles enter from the left side of the highway, among which HDVs exit from the right side of the highway, and CAVs need to cooperate highly to complete the driving task of the yellow cars exiting from the first ramp and the blue cars exiting from the second ramp. The main road of the highway has 3 lanes and the ramp has 1 lane.

\begin{table}
\centering
\caption{Simulation Environment Parameters.}
\label{table}
\setlength{\tabcolsep}{3pt}
\begin{tabular}{cc}
\toprule
\centering Parameters & \centering Value \tabularnewline
\hline
\centering Number of CAVs & \centering 4 \tabularnewline
\centering Number of HDVs & \centering 10 \tabularnewline
\centering $v_{max}$ & \centering 25 $m/s$ \tabularnewline
\centering $x_{road}$ & \centering 400 $m$ \tabularnewline
\centering HDVs departure speed & \centering 5 $m/s$ \tabularnewline
\centering CAVs departure speed & \centering 10 $m/s$ \tabularnewline
\centering Number of lanes & \centering 3 \tabularnewline
\centering First ramp exit location & \centering 250 $m$ \tabularnewline
\centering Second ramp exit location & \centering 370 $m$ \tabularnewline
\centering Simulation step & \centering 0.5 $s$ \tabularnewline
\bottomrule
\end{tabular}
\end{table}

We established a simulation environment for multi-vehicle collaborative decision-making training based on the driving environment, and deployed HDVs and CAVs according to the parameters in Table \uppercase\expandafter{\romannumeral1}. In the simulation, the lateral and longitudinal control modules of the vehicle are included. The longitudinal acceleration action of HDVs is generated by the Intelligent Driver Model (IDM) [29], the lateral lane change model is LC2013 [30], and the driving action instructions of CAVs are generated by the $Q$ value as mentioned before.

\subsection{Multi-vehicle Decision-making Progress}
As previously mentioned, the multi-vehicle collaborative decision-making process can be modeled as a MDP, which primarily consists of state representation, action space and reward function.

1) \textbf{State space} $s$: At any time $t$, the state space $s_t=[SR;N]$ contains two parts of information. The scene information, $SR$ represents the local scene occupancy grid constructed by each CAV in an agent-centric manner. The motion information, $N$ contains the motion features of all vehicles within the scene.

2) \textbf{Action space} $a$: In this study, we construct a discrete action space set with the aim that at each time step, all CAVs can learn lateral and longitudinal driving actions simultaneously. The lateral actions include changing lanes to the left, keeping lanes, and changing lanes to the right, and the longitudinal actions include accelerating, maintaining speed, and decelerating. Specifically, it can be expressed as follows:
\begin{equation}A=\{(a_{lc},a_{acc})|a_{lc} \in A_{lc}, a_{acc} \in A_{acc}\}\label{eq}\end{equation}
where $A_{lc}=\{LC,LK,RC\}$ and $a_{acc}=\{AC,MS,DC\}$.

3) \textbf{Reward function} $r$: In decision-making models based on deep reinforcement learning, the ultimate model performance hinges on the reward function design and the weight distribution of the rewards [31]. The design of the reward function in this paper strikes a balance among traffic efficiency, driving tasks and safety.

In order to enhance the traffic efficiency of CAVs, a reward function is designed based on the overall average speed, aiming to encourage high-speed driving, which is specifically expressed as follow:
\begin{equation}R_{speed}=\frac{1}{m}\sum_{i=1}^{m}\frac{v_i}{v_{max}}\label{eq}\end{equation}
where $v_i$ is the speed of each CAV, $v_{max}$ is the maximum speed limit of the highway, and $m$ is the number of all CAVs in the scene.

\begin{figure}[!t]
\centerline{\includegraphics[width=\columnwidth]{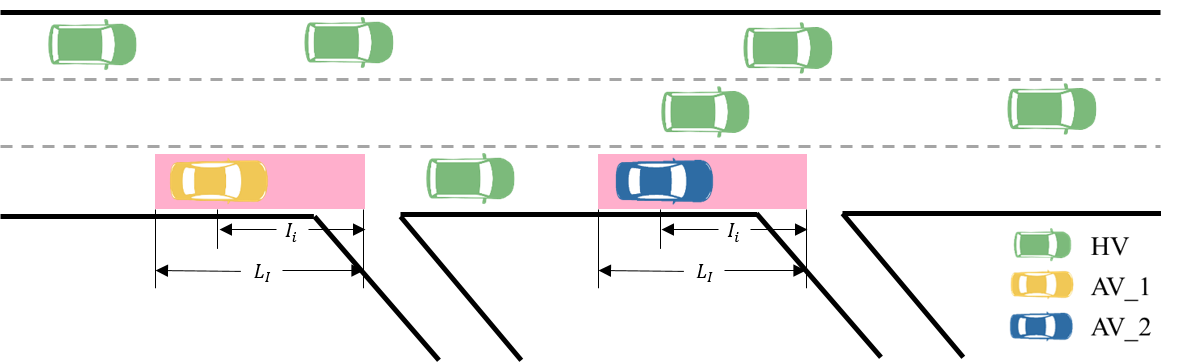}}
\caption{Intention reward diagram. When vehicles are about to off-ramp, we encourage them to drive to the right (pink area).}
\label{fig5}
\end{figure}

Another component of the reward function is the intention reward. We modify the practice of some other works [32], [33], [34] that is, only reward when the vehicle reaches the target, because this may cause CAVs to change lanes arbitrarily on the highway to reach the target, which will disrupt the entire mixed traffic. Specifically, the intention reward is when a CAV is about to leave the next ramp. We design a reward area in the rightmost lane 50 meters before the exit. The specific schematic diagram is shown in Fig. 5. When the CAV is driving in the reward area (pink area), the closer it is to the ramp exit, the higher the reward it will receive. The formula is as follows:
\begin{equation}R_{intention}=\sum_{i=1}^{m}(1-\frac{I_i}{L_I})\label{eq}\end{equation}
where $I_i$ represents the distance from the exit ramp, and $L_I$ represents the length of the reward area, which is 50 meters.

Collision penalty is the key to ensure that CAVs make safe decisions. We set $R_{collision}=-N_{collision}$, where $N_{collision}$ represents the number of collisions in each time step.

Finally, the entire reward function is expressed as follows:
\begin{equation}R=w_1R_{speed}+w_2R_{collision}+w_3R_{intention}\label{eq}\end{equation}
where $w_1$, $w_2$, $w_3$ are weight coefficients.

\subsection{Evaluating Indicator}
In order to evaluate the performance of GITSR in a driving environment, we collect task success rate, number of collisions, average speed and return as evaluation indicators during training, which are described as follows:

1) \textbf{Task success rate}:It quantifies the proportion of CAVs that successfully exit the designated ramp at the end of each training episode, reflecting the efficacy of the implemented training process strategy.

2) \textbf{Number of collisions}: It tallies the total number of collisions among all CAVs at the end of each training episode, reflecting the safety of the driving strategy.

3) \textbf{Average speed}: It calculates the average speed of all CAVs in the traffic flow at the end of each training episode, reflecting the efficiency of the driving strategy.

4) \textbf{Return}: The cumulative return of each training episode reflects the comprehensive performance of traffic efficiency, safety, and effectiveness of all CAVs cooperative driving strategies.

\subsection{Performance comparison and ablation experiments}
In order to evaluate the performance of the GITSR algorithm and the significance of the framework design, we conduct the following performance comparison and ablation experiments.

In our work, we use Multi-Agent Deep Q-Networks (MADQN) as the baseline algorithm, and also compare it with the Transformer encoding only (MADQN\_Transformer). In order to explore the role of agent-centric scene representation in multi-vehicle collaborative decision-making, we conduct the following two ablation experiments: 1) Comparing the results with and without scene representation in the MADQN algorithm; 2) Comparing the results of scene-centric direct input and agent-centric scene reconstruction in all algorithms.

\begin{table}
\centering
\caption{Parameters Used in Experiment.}
\label{table}
\setlength{\tabcolsep}{3pt}
\begin{tabular}{cc}
\toprule
\centering Parameters & \centering Value \tabularnewline
\midrule
\centering Transformer blocks & \centering 2 \tabularnewline
\centering Number of heads & \centering 4 \tabularnewline
\centering Embedding length & \centering 128 \tabularnewline
\centering Dimension of head & \centering 32 \tabularnewline
\centering Training episodes & \centering 3000 \tabularnewline
\centering Step size of warm-up & \centering 20000 \tabularnewline
\centering Batch size & \centering 32 \tabularnewline
\centering Replay buffer capacity & \centering 1e6 \tabularnewline
\centering Optimizer & \centering Adam \tabularnewline
\centering Discount factor & \centering 0.9 \tabularnewline
\centering $\epsilon$ decay steps & \centering 40000 \tabularnewline
\centering $\epsilon$& \centering 0.99 $\rightarrow$ 0.001 \tabularnewline
\centering $w_1$, $w_2$, $w_3$& \centering 3, 9, 15 \tabularnewline
\bottomrule
\end{tabular}
\end{table}

\subsection{Implementation details}
The relevant parameters of our experiment are shown in Table \uppercase\expandafter{\romannumeral2}. The total number of training episodes is 3000. A warm-up phase of 20,000 steps is set before training. CAVs randomly execute actions and store them in the replay pool, which is defined as $\pi(s)=\text{random}(a)$. During the training phase, CAVs make decisions based on $Q$ values and the $\epsilon$ exploration strategy. The $\epsilon$ exploration strategy is that when CAVs make decisions at each step, there is an $\epsilon$ probability to execute random actions, and a $1-\epsilon$ probability to select a strategy based on the $Q$ value. The specific formula is:
\begin{equation}
\pi(s) = 
\begin{cases} 
\text{random}(a) & \text{P} = \epsilon \\ 
\arg\max Q(s, a) & \text{P} = 1 - \epsilon 
\end{cases}
.\label{eq}
\end{equation}

We use Adam optimizer in Pytorch to train the model with a learning rate of 1e-4. All methods are trained three times with random seeds, and each training takes about 6 hours on an Intel Core i9-10920 CPU and an NVIDIA GeForce RTX 3090 GPU.

\section{Result and discussions}
\label{sec:result and discussions}

This section will present and analyze our experimental results, including comparisons with baseline methods and ablation experiments.

Figure 6 shows the performance comparison between our method and the baseline methods. Overall performance from the reward return during the training process, it can be seen that the performance of GITSR is significantly better than MADQN\_Transformer and MADQN, proving the effectiveness of the GITSR framework. The results of the task success rate show that it is necessary to model spatial interaction behaviors through GNN in the GITSR framework, which can improve the stability of multi-vehicle collaborative strategies. At the same time, we use the number of collisions in each episode to evaluate the safety of the algorithm, which shows that the self-attention mechanism of the Transformer encoder can help each CAV capture the traffic scene information around the vehicle and help CAVs make safe decisions. We hope that all CAVs can improve the overall efficiency of traffic flow while completing the driving task collaboratively. 
Fig. 7 shows the overall average speed of CAVs during the training process. GITSR achieves a good balance between safety and efficiency. The driving behavior of MADQN\_Transformer is more conservative, resulting in slower speed, while MADQN sacrifices safety to maintain the highest speed, which is not conducive to autonomous driving decisions. In summary, GITSR shows better performance in many aspects compared with the baseline methods.

\begin{figure*}[!t]
\centerline{\includegraphics[width=\textwidth]{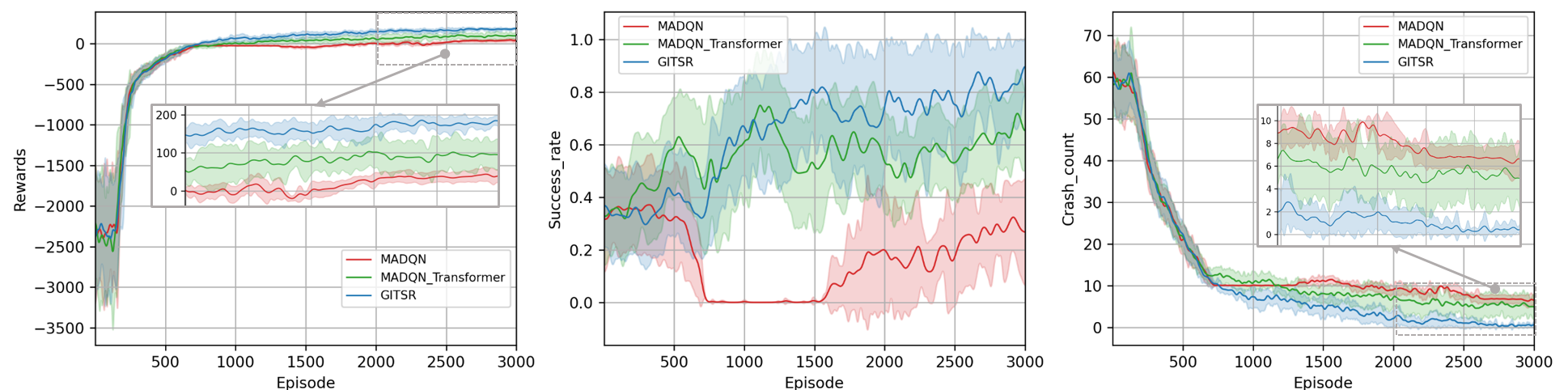}}
\caption{Evaluation index curves of GITSR and baseline methods during training, from left to right, are reward return, task success rate, and number of collisions. The experimental results are drawn under three random seeds.}
\label{fig6}
\end{figure*}

\begin{figure}[!t]
\centerline{\includegraphics[width=\columnwidth]{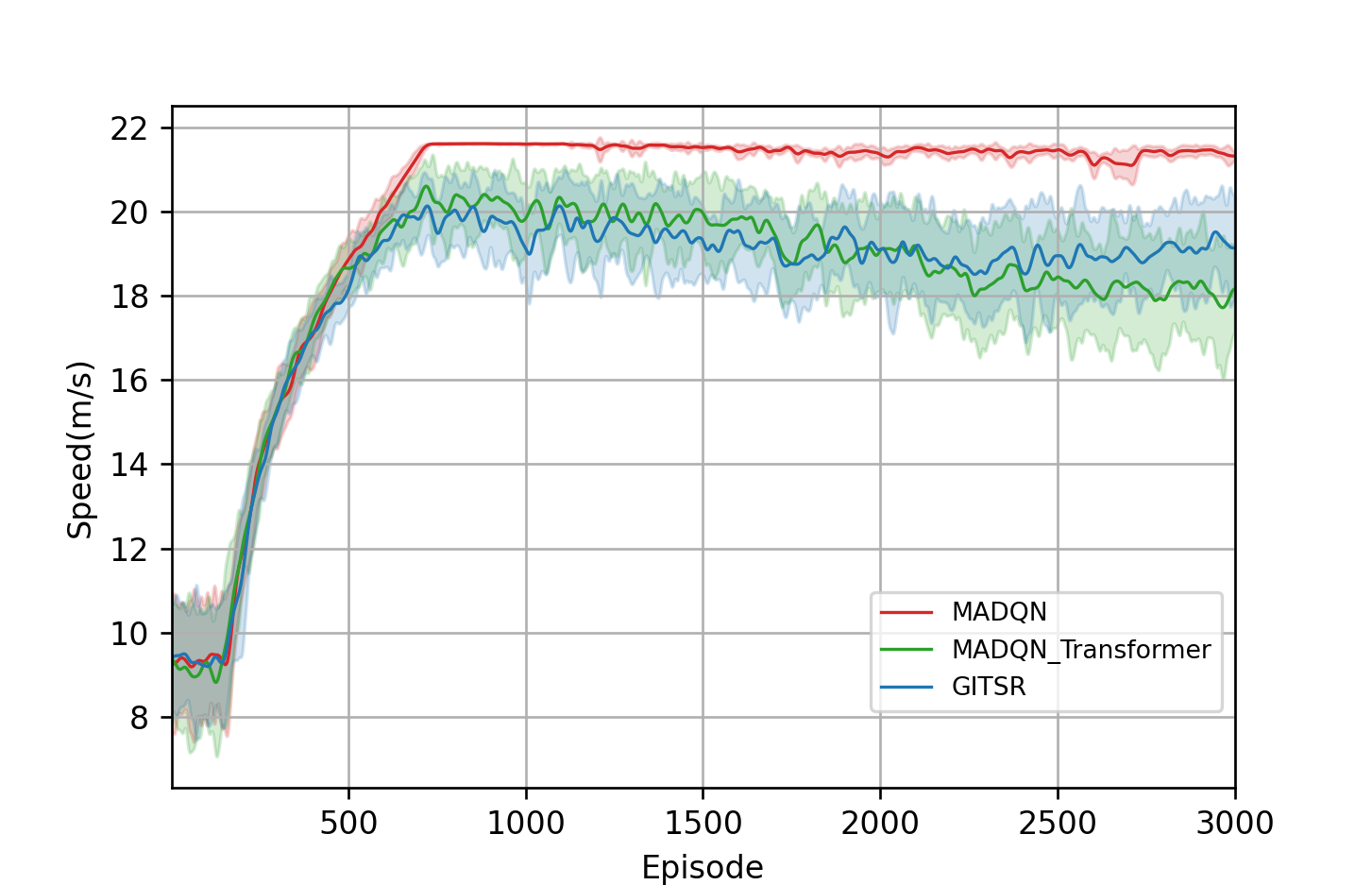}}
\caption{The overall average speed of CAVs during the training process of different algorithms under three random seeds.}
\label{fig7}
\end{figure}

%Fig. 8 shows the results of the ablation experiment comparing the presence and absence of scene representation in the MADQN algorithm. Experimental results show that in addition to the motion information of traffic participants in the driving scene, it is necessary to extract the scene representation information of CAVs from the scene. Although there is no significant improvement in the mission success rate, it effectively reduces the number of collisions, indicating that scene representation can help CAVs better understand scene information and make safer decisions. Adding scene representation in MADQN brings higher computational burden and inference difficulty to the model, and therefore is slower in convergence speed than methods without scene representation.

%\begin{figure*}[!t]
%\centerline{\includegraphics[width=\textwidth]{fig8.png}}
%\caption{Ablation experiment with and without scene representation under the MADQN algorithm: the green curve represents the state representation result with only the feature list, and the red curve represents the result with the scene representation added.}
%\label{fig8}
%\end{figure*}

\begin{figure*}[!t]
\centerline{\includegraphics[width=\textwidth]{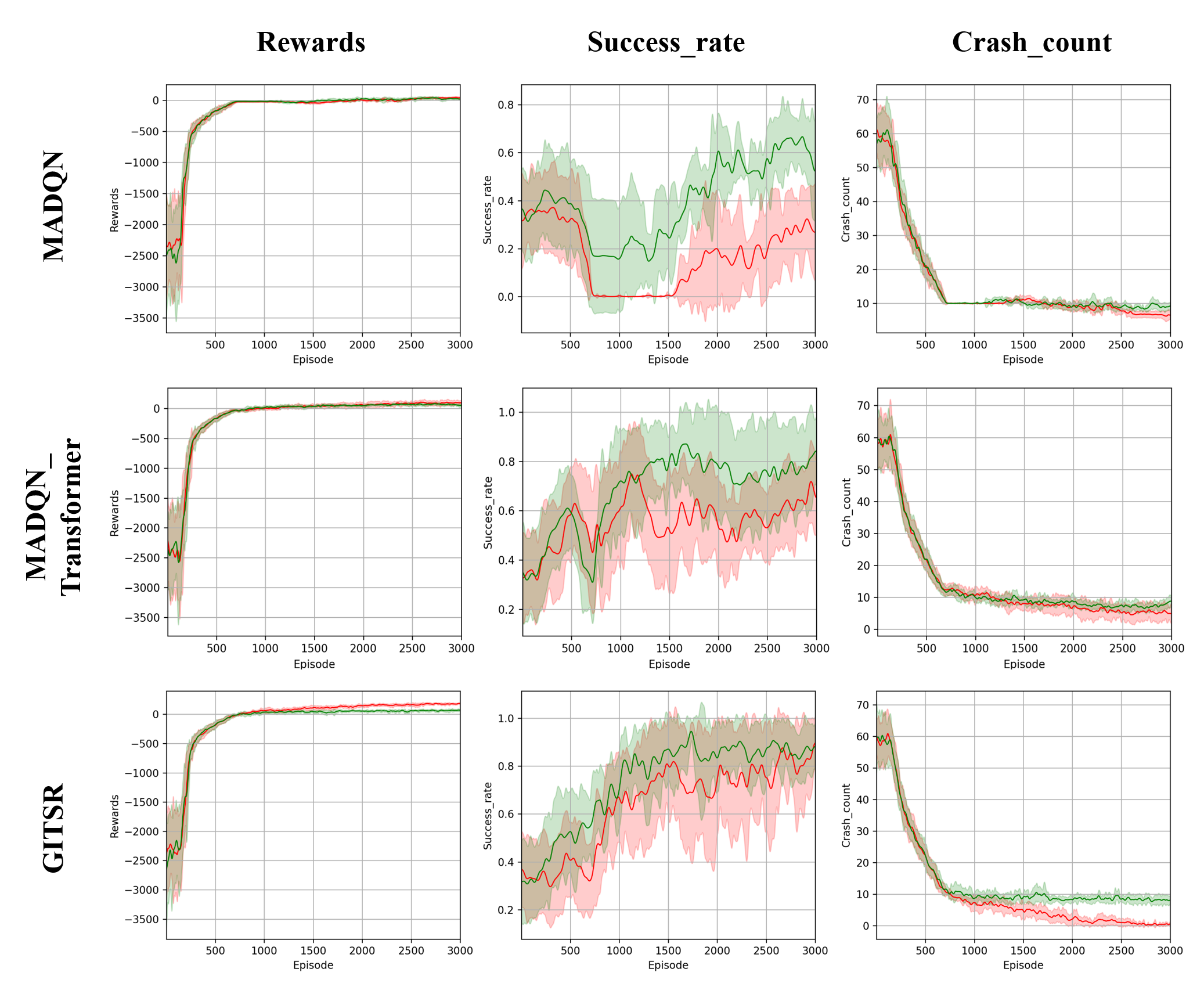}}
\caption{The effects of the three algorithms under scene-centric and agent-centric scene representation are compared. The agent-centric model achieves better results, but scene-centric has more advantages in terms of convergence speed and computational burden.}
\label{fig9}
\end{figure*}

We are pleasantly surprised by the ablation experiment results comparing scene-centric direct input and agent-centric scene reconstruction among all algorithms, as shown in Fig. 8. The experimental results show that the agent-centric scene reconstruction method can help CAVs understand the surrounding traffic scenes more easily and effectively reduce the number of collisions. However, in terms of task success rate, scene-centric shows better performance. We speculate that because scene-centric does not need to re-infer the local traffic scenes of CAVs and uses a fixed coordinate system, it can directly obtain the target point of task completion from map features, which is interesting for downstream planning tasks. In addition, the scene-centric method has a smaller computational burden, while agent-centric needs to model all CAVs, which will be a computational bottleneck for large traffic scenes.

\section{Conclusion}
In this study, we introduce GITSR, an effective graph interaction Transformer-based scene representation reinforcement learning framework for improving collaborative decision-making of autonomous vehicles. The framework mainly includes: Transformer is used to encode agent-centric local scene input to capture interactive information of surrounding traffic scenes; GNN is used to refine the motion information of traffic participants to represent the spatial interaction characteristics of dynamic traffic scenes. Reinforcement learning algorithm MADQN splices the two parts of information as decision input and outputs collaborative driving behaviors. We verify it in a challenging interactive collaborative driving environment, and the results show that our method performs better than the baseline methods. We also study the performance and impact of different modules in GITSR and find that scene representation can help CAVs better understand the scene and effectively reduce the number of collisions. The scene-centric scene representation has a higher task success rate, while the agent-centric scene representation is better in terms of safety.

It is undeniable that although the current algorithm has achieved excellent performance in multi-vehicle collaborative decision-making, the increase in the number of CAVs will inevitably bring a higher secondary modeling burden in scene representation, which is not conducive to large-scale intelligent transportation. We believe that the efficient reasoning speed of scene-centric scene representation and its independence from the number of CAVs can enable better performance in large-scale scenes, especially in planning tasks. In future work, we will focus on exploring more effective scene representation framework in large-scale scenarios, improving the understanding of dynamic scenes and collaborative driving decision-making capabilities of CAVs.

\printbibliography[heading=bibintoc, title=\ebibname]

\end{document}